\documentclass[conference]{IEEEtran}
\IEEEoverridecommandlockouts

\def\BibTeX{{\rm B\kern-.05em{\sc i\kern-.025em b}\kern-.08em
    T\kern-.1667em\lower.7ex\hbox{E}\kern-.125emX}}
\usepackage{comment}
\usepackage{cite}
\usepackage{amsmath,amssymb,amsfonts}
\usepackage{graphicx}
\usepackage{textcomp}
\usepackage[table]{xcolor}
\usepackage{ragged2e}
\usepackage{placeins}
\usepackage{comment}
\usepackage{pgfgantt}

\usepackage{lipsum}
\usepackage{multicol}
\usepackage{url}
\usepackage[linesnumbered,ruled]{algorithm2e}
\usepackage{algpseudocode}

\usepackage{tikz}
\usetikzlibrary{shapes.geometric,arrows.meta,positioning,decorations.markings,decorations.pathreplacing}
\tikzstyle{my right of} = [right=of #1.east]
\tikzstyle{my left of} = [left=of #1.west]

\newcommand\copyrighttext{%
  \footnotesize \textbf{To appear in IEEE AITesting 2020} \\
  \footnotesize \textcopyright 2020 IEEE. Personal use of this material is permitted. Permission from IEEE must be obtained for all other uses, in any current or future 
  media, including reprinting/republishing this material for advertising or promotional purposes, creating new collective works, for resale or redistribution to servers or lists, or reuse of any copyrighted component of this work in other works.}
\newcommand\copyrightnotice{%
\begin{tikzpicture}[remember picture,overlay]
\node[anchor=south,yshift=10pt] at (current page.south) {\fbox{\parbox{\dimexpr\textwidth-\fboxsep-\fboxrule\relax}{\copyrighttext}}};
\end{tikzpicture}%
}

\usepackage{subcaption}

\begin{document}
% The best title in the world 
\title{Controlled time series generation for automotive software-in-the-loop testing using GANs}

\author{
% Author list AND affiliations inside 
\IEEEauthorblockN{Dhasarathy Parthasarathy\IEEEauthorrefmark{2}\IEEEauthorrefmark{1}, Karl B\"ackstr\"om\IEEEauthorrefmark{1}, Jens Henriksson\IEEEauthorrefmark{1}\IEEEauthorrefmark{4}, S\'olr\'un Einarsd\'ottir\IEEEauthorrefmark{1}} \\
\IEEEauthorblockA{\IEEEauthorrefmark{2}Volvo Group, Gothenburg, Sweden, Email: dhasarathy.parthasarathy@volvo.com}
\IEEEauthorblockA{\IEEEauthorrefmark{1}Chalmers University of Technology, Gothenburg, Sweden, Email: \{bakarl, slrn\}@chalmers.se}
\IEEEauthorblockA{\IEEEauthorrefmark{4}Semcon AB, Gothenburg, Sweden, Email: jens.henriksson@semcon.com}
}

\maketitle
\copyrightnotice

\begin{abstract}
Testing automotive mechatronic systems partly uses the software-in-the-loop approach, where systematically covering inputs of the system-under-test remains a major challenge. In current practice, there are two major techniques of input stimulation. One approach is to craft input sequences which eases control and feedback of the test process but falls short of exposing the system to realistic scenarios. The other is to replay sequences recorded from field operations which accounts for reality but requires collecting a well-labeled dataset of sufficient capacity for widespread use, which is expensive. This work applies the well-known unsupervised learning framework of Generative Adversarial Networks (GAN) to learn an unlabeled dataset of recorded in-vehicle signals and uses it for generation of synthetic input stimuli. Additionally, a metric-based linear interpolation algorithm is demonstrated, which guarantees that generated stimuli follow a customizable similarity relationship with specified references. This combination of techniques enables controlled generation of a rich range of meaningful and realistic input patterns, improving virtual test coverage and reducing the need for expensive field tests.
\end{abstract}


%%%%%%%%%%%%%%%%%%%%%%%%%%%%%%%%%%%%%%%%%%%%%%%%%%%%%%%%%%%%%%%%%%%%%%%%%%%%%%%%
\section{Introduction}
\label{sec:intro}
Software-in-the-loop (SIL) is a commonly used technique for developing and testing automotive application software. In a typical SIL setup, the System Under Test (SUT) is a combination of the application software and software representations of related physical and virtual entities. By executing the mechatronic SUT in a purely virtual environment, SIL testing enables continuous integration and verification with faster test feedback~\cite{kaijser:hal-02156371}. The extent and fidelity of virtual execution are usually defined by the test objective. This work focuses on one important component of the system environment -- input stimuli. Input stimulation in a SIL setup can be challenging since ranges are often large and continuous, with developers/testers aiming for systematic coverage, while trading off concerns such as cost and duration of testing, coherence of test results, and confidence in covering corner cases. 

\subsection{Challenges in current practice}
\begin{figure}[h]
    \centering
    \begin{subfigure}{\linewidth}
        \includegraphics[width=0.98\linewidth]{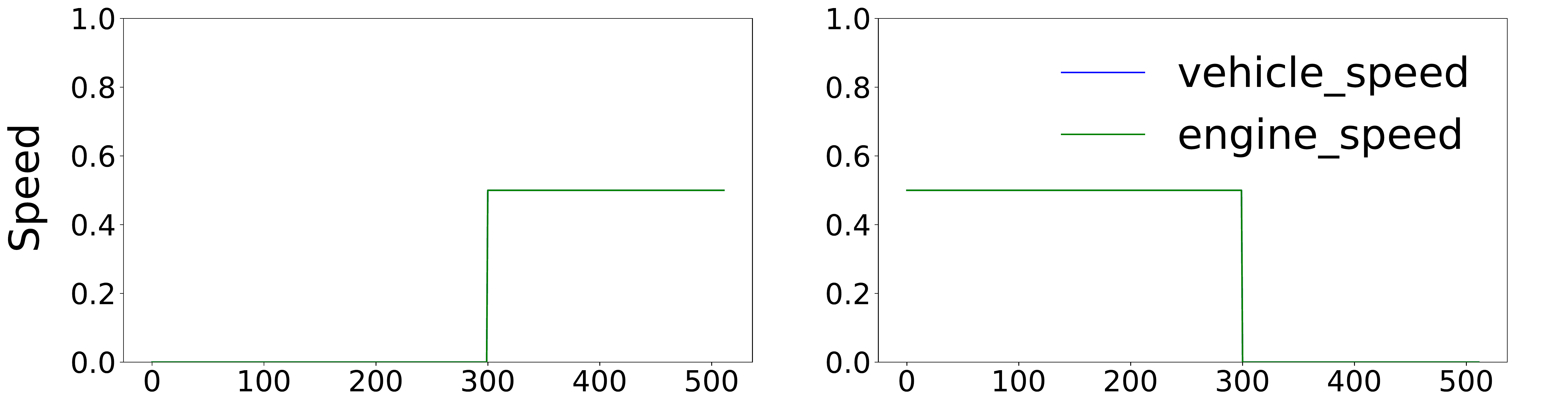}
        \caption{Crafted time series}
        \label{fig:ref-signals-tmpl}
    \end{subfigure}
    \begin{subfigure}{\linewidth}
        \includegraphics[width=0.98\linewidth]{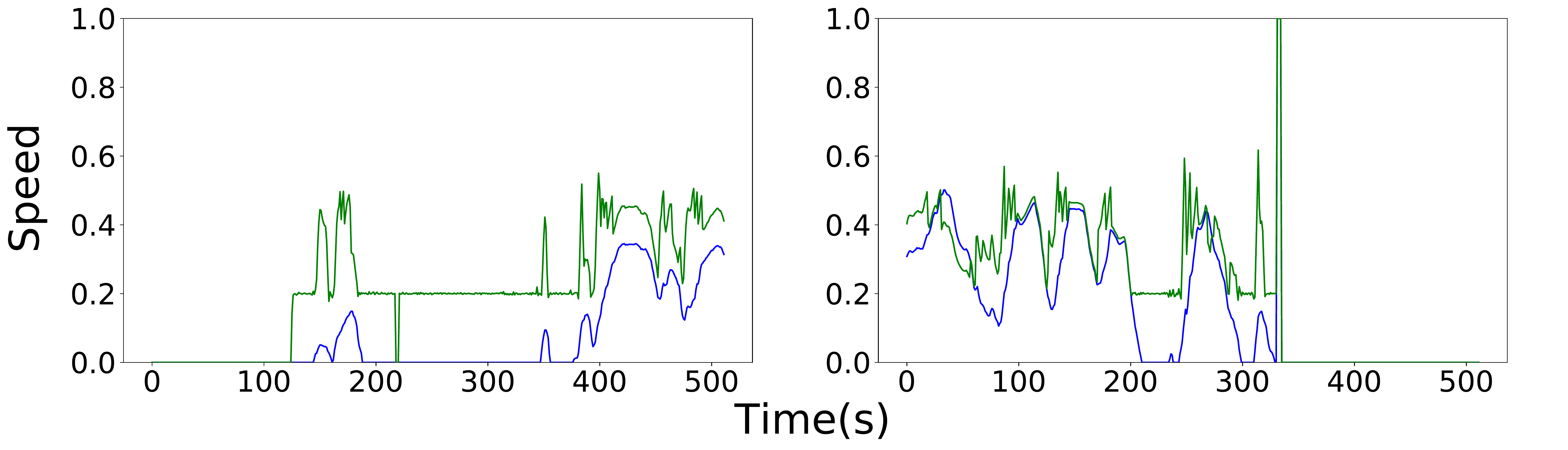}
        \caption{Recorded time series}
        \label{fig:ref-signals-real}
    \end{subfigure}
    %reduce space after caption
    %\setlength{\belowcaptionskip}{-10pt}
    \caption{Two approaches of input stimulation (speed values rescaled to  [0,1])}
\end{figure}

One technique for sampling the input space is to craft templates, designed to capture characteristics that are interesting for the test. The Worldwide harmonized Light-duty vehicles Test Cycles (WLTC)\cite{wltp}, for example, are designed to represent driving cycles (time series of vehicle and/or engine speed) under typical driving conditions, and are popular for benchmarking drive train applications. Standard templates apart, it is common practice for a tester to craft custom templates for a test case. Templates, as seen in Fig~\ref{fig:ref-signals-tmpl} of a vehicle starting and stopping, enable testers to perform controllable and repeatable tests. When templates need to be more realistic, testers spend significant effort in finding the right function approximations. Templates, no matter how complex, bear the risk of not covering realistic scenarios experienced in the field. \looseness=-1

To better approximate reality, a common practice in the automotive industry is to apply real-world input stimuli that are recorded from vehicles in the field. Examples can be seen in Fig~\ref{fig:ref-signals-real}, which show recorded sequences of a vehicle starting and stopping. To control the test process, testers need to identify those exact sequences which possess desired characteristics. A special case, fault tracing, occurs when a tester chooses sequences during which faults are known (or suspected) to have occurred. To test using the right field-recorded sequences, one needs a manageable data set of labeled sequences to choose from, which is an expensive requirement. Additionally, literal signal replay, while helpful in recreating field scenarios, has the limitation of being largely restricted to what is recorded. In practice, finding the right balance between the two techniques of input stimulation is a perennial concern. This limits applicability and the overall confidence in SIL testing.\looseness=-1

\subsection{Contributions}
While drawbacks do exist in each approach, it is easy to see that both real-world and designed stimuli have good utility. Bridging between the two approaches, this work demonstrates a method for controlled generation of realistic synthetic stimuli. Specifically, we contribute:
\begin{itemize}
    \item A Variational Autoencoder-Generative Adversarial Network (VAE/GAN) model for producing synthetic vehicle signal sequences for SIL testing
    \item A method to generate realistic SIL stimuli that are measurably similar to reference stimuli, improving chances of discovering bugs
    \item A method to customize stimulus generation using similarity metrics
\end{itemize}

Together, these techniques give better control in covering the input space and ease the test process. 

\subsection{Scope}
While the focus is on controlled generation of time series of in-vehicle signals (information exchanged between on-board applications), our results may be applicable to time series generation in other domains. With signals being both discrete and continuous-valued, the scope of this work is limited to the latter, since covering continuous input ranges is a pressing challenge in SIL testing.

\subsection{Structure}
%\begin{comment}
Forthcoming sections present a survey of related techniques (Section~\ref{sec:rw}), followed by an introduction of the data and model architecture (Section~\ref{sec:model}). Then comes the presentation on using similarity metrics in model training and evaluation (Section~\ref{sec:training}), and in controlled generation of synthetic stimuli (Section~\ref{sec:generation}). This is followed by a discussion on the idea of similarity with references as a plausibility measure (Section~\ref{sec:similarity}), and the conclusion (Section~\ref{sec:conc}).
%\end{comment}
%%%%%%%%%%%%%%%%%%%%%%%%%%%%%%%%%%%%%%%%%%%%%%%%%%%%%%%%%%%%%%%%%%%%%%%%%
\section{Related work}
\label{sec:rw}
Following its introduction \cite{DBLP:conf/nips/GoodfellowPMXWOCB14},  GANs have been applied for synthetic data generation in a variety of domains. Reflecting our scope, we focus on surveying reports of continuous-valued sequence generation and highlighting the following aspects -- (i) the model architecture and (ii) measures to check plausibility of generated samples.

Using a recurrent GAN, generation of real-valued medical time-series was demonstrated by \cite{DBLP:journals/corr/EstebanHR17}, where two plausibility measures were shown. One is measuring Maximum Mean Discrepancy (MMD) \cite{Sriperumbudur:2010:HSE:1756006.1859901} between populations of real and synthetic samples, and the other to train a separate model using synthetic/real data and evaluate it using real/synthetic data. Using a recurrent architecture, \cite{DBLP:journals/corr/Mogren16} demonstrated music generation, with generated samples evaluated using observable features such as polyphony and repetitions. From an application perspective, \cite{zec:rcgan-sensors} comes close, where a recurrent GAN was applied to generate time series of automotive perception sensors, for simulation-based verification. Plausibility is shown using the Jensen-Shannon (JS) divergence, as a measure of symmetric relative entropy between populations of real and generated samples. A combined Long-Short Term Memory (LSTM) and Mixture Density Network (MDN) GAN for generating sensor data has been shown by~\cite{DBLP:journals/corr/AlzantotCS17}, where GAN loss and discriminator prediction are used as evaluation measures. Convolutional GANs have been applied for sequence generation in \cite{DBLP:journals/corr/abs-1902-05624} and \cite{DBLP:conf/smartgridcomm/ZhangKKP18}. Both of them apply MMD and Wasserstein-1 \cite{7974883}, while the latter additionally applies classical machine learning methods like k-means clustering to measure plausibility. Beyond deep generative models, purely statistical methods of generating and evaluating time series (examples \cite{DBLP:journals/corr/abs-1903-02787} and \cite{DBLP:conf/ssdbm/KegelHL18}) have been reported. However, such methods are, in the existing literature, inherently dependent on features selected by a domain expert. In comparison, a \textit{deep learning} or \textit{GAN} approach provides the capability to learn the necessary features automatically, thus being able to optimally adapt to the dataset on which it is applied.

The SIL testing process normally involves test case design, where the right (class of) input stimuli, \textit{the reference}, are chosen or constructed to meet the test objective. Therefore, unlike most reported applications, to maximize confidence, SIL stimulus generation must verifiably comply with the reference. We show how such compliance can be guaranteed using a novel combination of network architecture and metric-based search.\looseness=-1
%%%%%%%%%%%%%%%%%%%%%%%%%%%%%%%%%%%%%%%%%%%%%%%%%%%%%%%%%%%%%%%%%%%%%%%%%%%%%%%%%%%%%%%%%%%%%%%%%%%%%%%%%%%%%%%%%%%%%%%%%%%%%%%%%%%%%%%
\section{Model setup}
\label{sec:model}
The following sections describe the chosen model architecture and dataset. All code and data used in this work is publicly available\footnote{\url{https://github.com/dhas/LoGAN.git}}.\looseness=-1

\subsection{Choosing the VAE/GAN architecture}
Generative models learn to encode information about a distribution of samples $x$ in the form of a concise representation. The GAN uses a \textit{representation $\xrightarrow{}$ generator $\xrightarrow{}$ discriminator} architecture. Given the latent representation $z \sim p(z)$, a generator network maps it to a data sample $\bar{x} \sim Gen(z)$, whose membership in the sample distribution is assessed by the discriminator network $Disc(\bar{x})$. Training the GAN involves playing the minimax game, minimizing the objective in Eq~\ref{eq:gan}\cite{DBLP:conf/nips/GoodfellowPMXWOCB14}. The generator tries to fool the discriminator with synthetic samples, while the discriminator criticizes generated samples.\looseness=-1

\begin{equation}
    \mathcal{L}^{*}_{GAN}=\log(Disc(x)) + \log(1-Disc(Gen(z))) \label{eq:gan} \\
\end{equation}

The alternative Variational Autoencoder (VAE) uses the \textit{encoder $\xrightarrow{}$ representation $\xrightarrow{}$ decoder} architecture. Given a sample $x$, it uses an encoder network to map it to a latent representation $z \sim q(z|x) = Enc(x)$ and a decoder network to reconstruct an estimate of the original sample $\bar{x} \sim p(x|z) = Dec(z)$. Training the VAE involves minimizing element-wise reconstruction error between $x$ and $\bar{x}$ (Eq~\ref{eq:pix-met}), while jointly using a regularization term (Eq~\ref{eq:prior}) \cite{DBLP:journals/corr/KingmaW13} that encourages smoother encoding of the latent space.

\begin{gather}
    \mathcal{L}_{elem} = -\mathbb{E}_{q(z|x)}[\log \bar{x}] \label{eq:pix-met} \\
    \mathcal{L}_{prior} = -\mathbb{E}_{q(z|x)}[\log(p(x|z)] \label{eq:prior}
\end{gather}

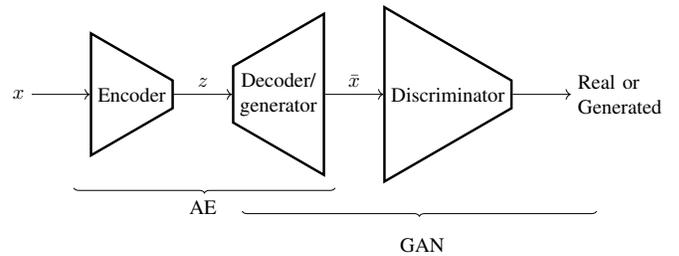
\begin{figure}[t]
% centering/sizing stuff here
\centering
\resizebox{9cm}{!}{
\begin{tikzpicture}[
      line/.style={thick},
      network/.style={draw, trapezium, line width=1.2pt},
      arrow/.style={thick, ->}
    ]
    % Draw the three network elements
    \node[network, shape border rotate=270] (encoder) {Encoder};
    \node[network, shape border rotate=90, my right of=encoder, align=center] (decoder) {Decoder/\\generator};
    \node[network, shape border rotate=270, my right of=decoder] (discriminator) {Discriminator};
    
    % Input/output
    \node[auto,my left of = encoder] (x) {\(x\)};
    \node[auto, align=left, my right of = discriminator] (Real/Gen) {Real or\\Generated};
    
    % Draw transitions
    \draw[->] (x) edge (encoder)
              (encoder) edge node[auto] (z) {\(z\)} (decoder)
              (decoder) edge node[auto] (xbar) {\(\bar{x}\)} (discriminator)
              (discriminator) edge (Real/Gen);
    ;
    
    % Draw the labels at the bottom
    \node[auto,below=1.7cm of z] (AE) {AE};
    \node[auto] at (5,-2.6) (GAN) {GAN};
    \draw[decorate, decoration={brace,mirror,raise=6pt}]
         (-1,-1.4) -- (3.5,-1.4);
    \draw[decorate, decoration={brace,mirror,raise=6pt}]
         (1.9,-1.8) -- (8,-1.8);

\end{tikzpicture}
}
    \caption{The VAE/GAN model}
    \label{fig:vae-gan}
\end{figure}

In any case, synthetic samples are generated by sampling $z$ and mapping it to $\bar{x}$. As noted earlier, in many SIL scenarios, testers are interested in stimuli that are similar to a reference, whether designed or field-recorded. Having a reference stimulus available as the starting point, an architecture with an encoder readily locates it in the latent space, easing the sampling process. While VAE seems suitable, it has well-recognized drawbacks in using element-wise reconstruction loss, which results in generated samples of poorer quality compared to those generated by a GAN. One alternative is the VAE/GAN model, which combines both frameworks and has an \textit{encoder~$\xrightarrow{}$ representation~$\xrightarrow{}$ generator~$\xrightarrow{}$ discriminator} architecture (Fig~\ref{fig:vae-gan}). Since the discriminator learns the essential features of a convincing sample, a better measure of reconstruction would be the distance between real and generated samples at the $l^{th}$-layer of the discriminator (Eq \ref{eq:sim-metric}). By minimizing ${L}_{Dis_l}$ instead of element-wise reconstruction, and by additionally discriminating reconstructed samples (Eq~\ref{eq:lgan}), the VAE/GAN has been shown\cite{DBLP:conf/icml/LarsenSLW16} to generate samples of better quality than VAE. With the double advantage of being able to encode the reference and generate samples of good quality, VAE/GAN is a good starting point for SIL stimulus generation. \looseness=-1

\begin{gather}
    \mathcal{L}_{Dis_l} = -\mathbb{E}_{q(z|x)}[\log p(Disc_l(x)|z)] \label{eq:sim-metric}\\
    \mathcal{L}_{GAN} = \mathcal{L}^{*}_{GAN} + \log(1-Disc(Dec(Enc(x)))) \label{eq:lgan} \\
    \mathcal{L} = \beta \mathcal{L}_{prior} + \gamma \mathcal{L}_{Dis_l} + \mathcal{L}_{GAN} \label{eq:vaegan}
\end{gather}

\subsection{The data set}
To illustrate our technique of controlled generation, fixed-length time series of two signals of interest, \textit{vehicle and engine speed}, were chosen. While any other set of signals can conceivably be chosen, the chosen set of two is a good starting point since they influence a broad range of on-board applications. The primary source of data is a set of signals, recorded in a fleet of 19 Volvo buses over a 3--5 year period. From this source, 100k 512--sample long unlabeled sequences of the signals of interest were collected to form the training set. Fixing the length and aligning the sequences helps treating them like images, allowing easy use of standard convolutional GAN architectures. An additional test set of 10k samples was collected. In this test set, a sparse labeling activity was conducted to pick sequences with desired characteristics. Examples of identified start and stop sequences were already seen in Fig~\ref{fig:ref-signals-real}. More samples from the test set can be seen in Fig~\ref{fig:test-set} in the appendix.\looseness=-1

\subsection{Designing the model}
Having fixed the sequence length, upon trial-and-error in architecture selection, the encoder, decoder/generator, and discriminator are designed as 4--layer 1D convolutional neural networks. The kernel and mini-batch sizes, both influential hyper-parameters, are fixed to 8 and 128 respectively. Closely following the original VAE/GAN objective, the composite network is trained by minimizing the triple criteria set by Eq~\ref{eq:vaegan}. Two additional hyperparameters are introduced, $\beta$ following the $\beta$-VAE model, which encourages disentanglement \cite{DBLP:conf/iclr/HigginsMPBGBML17} between latent variables, and $\gamma$, which, as suggested in \cite{DBLP:conf/icml/LarsenSLW16}, allows weighing the influence of reconstruction against discrimination. Among suggestions in \cite{DBLP:journals/corr/RadfordMC15} to improve stability of GAN training, using the Adam optimizer with a learning rate of $2\cdot10^{-4}$ and a momentum of $0.5$ greatly helped in achieving training convergence, while the recommendation of using $LeakyReLU$ and $tanh$ activations did not. Based on visual inspection, $ReLU$ and $sigmoid$ activations are found to produce samples of better quality, perhaps because of a significant amount of baseline-zero values in recorded signals, when the buses were turned off.\looseness=-1
%%%%%%%%%%%%%%%%%%%%%%%%%%%%%%%%%%%%%%%%%%%%%%%%%%%%%%%%%%%%%
\section{Metric guided training and evaluation}
\label{sec:training}
GAN training is widely acknowledged as challenging, with $Gen$ and $Disc$ loss trends being the standard indicator of training progress. Tracking objectives during training does provide valuable feedback on model design and parameter selection. However, when it comes to assessing the quality of generated data, they are indirect at best. Testers, who are mainly interested in assessing the quality of generated stimuli, are better advised by direct measures. While visual inspection is common in the image domain, a quantitative alternative is the inception score \cite{NIPS2016_6125}, where generated images are evaluated by the widely-benchmarked Inception v3 model. In the absence of such a benchmark for in-vehicle signal time series, or perhaps even time series in general, alternative metrics are necessary.\looseness=-1

\subsection{Choosing metrics for evaluation}
As described earlier, setting up a SIL test involves choosing right stimulus patterns. Having picked a reference pattern, testers normally vary it in some predictable fashion, with variations expected to be structurally related to the reference. Combinatorially designing these variations, which is common standard practice, may make it easy to enforce the structural relationship, but can be overwhelming as sequences grow longer. Generated stimuli do eliminate the need for such manual design, but they remain useful only as long as they show a structural relationship with the reference. To enforce this, one can turn to a number of similarity measures, some of which were mentioned in Section~\ref{sec:rw}. However, with references simply being a handful of sequences (the fewer the easier for test design), metrics that specialize in measuring pair-wise similarity are more intuitive. Specifically, we consider:

\begin{itemize}
    \item Dynamic Time Warping (DTW) \cite{1163055} -- which is an objective measure of similarity between two sequences. DTW is a measure of distance and therefore a non-negative real number. Lower values indicate higher similarity.
    \item Structural Similarity Index (SSIM) \cite{1284395} -- which is an objective measure of similarity between two images. SSIM between a pair of images is a real number between 0 (dissimilar) to 1 (similar)
\end{itemize}

With our sequences being time series, DTW is perhaps a natural fit. SSIM may consider its input signals to be images, but it does not exclude itself from comparing aligned sequences of fixed length. It also has the advantage of being a number in the range $(0,1)$, making it easy to inspect. More important to note is that testers need to choose the right metric that helps meet the test objective.\looseness=-1

\subsection{Calibrating metrics during training}

\begin{figure}[b!]
    \begin{subfigure}{\linewidth}
        \centering
        \includegraphics[width=0.8\linewidth,trim={0 0 0 8},clip]{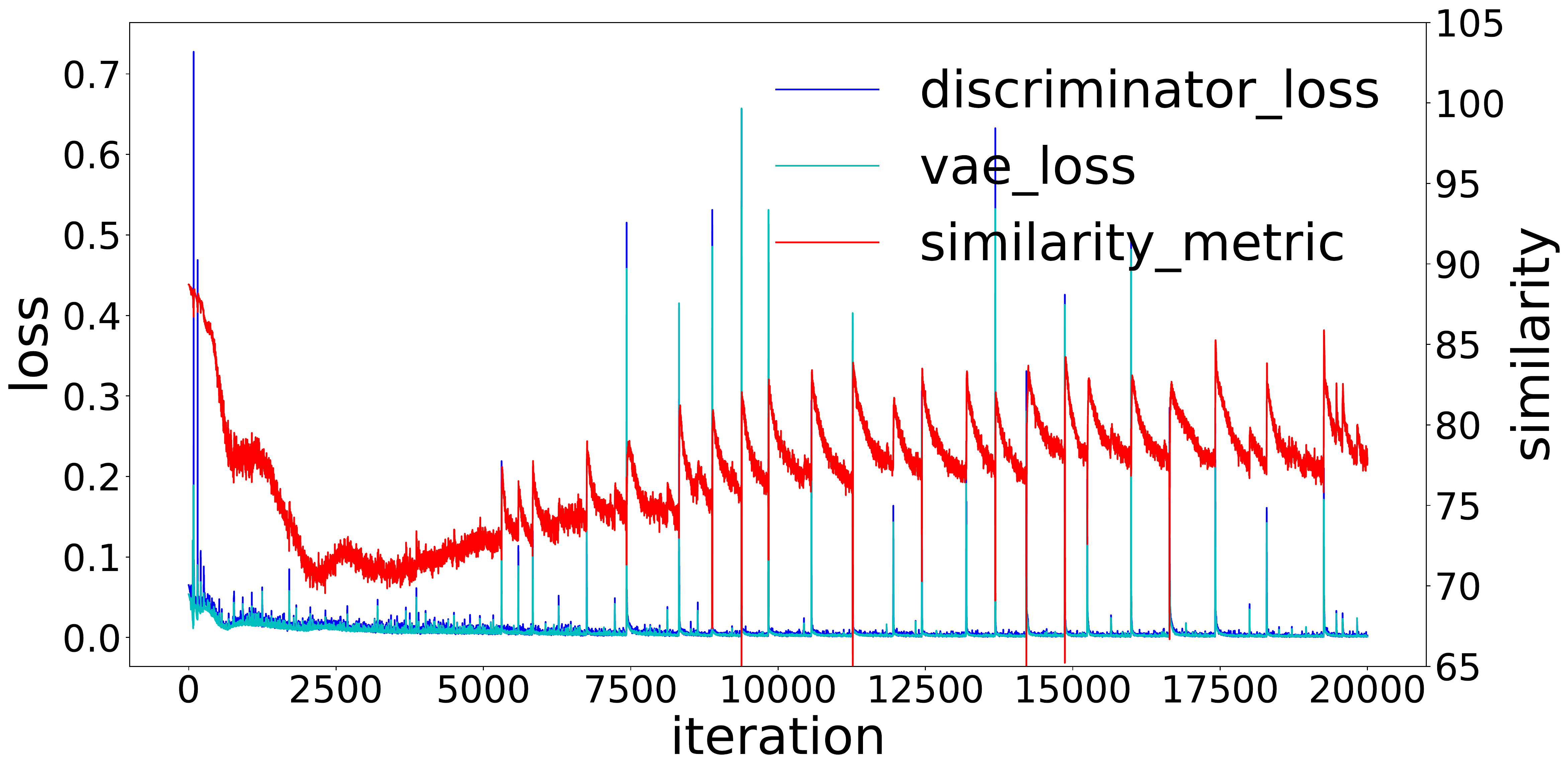}
        \caption{Training history}
        \label{fig:training-hist}
    \end{subfigure}
    \begin{subfigure}{\linewidth}
        \centering
        \includegraphics[width=0.8\linewidth,trim={0 0 0 8},clip]{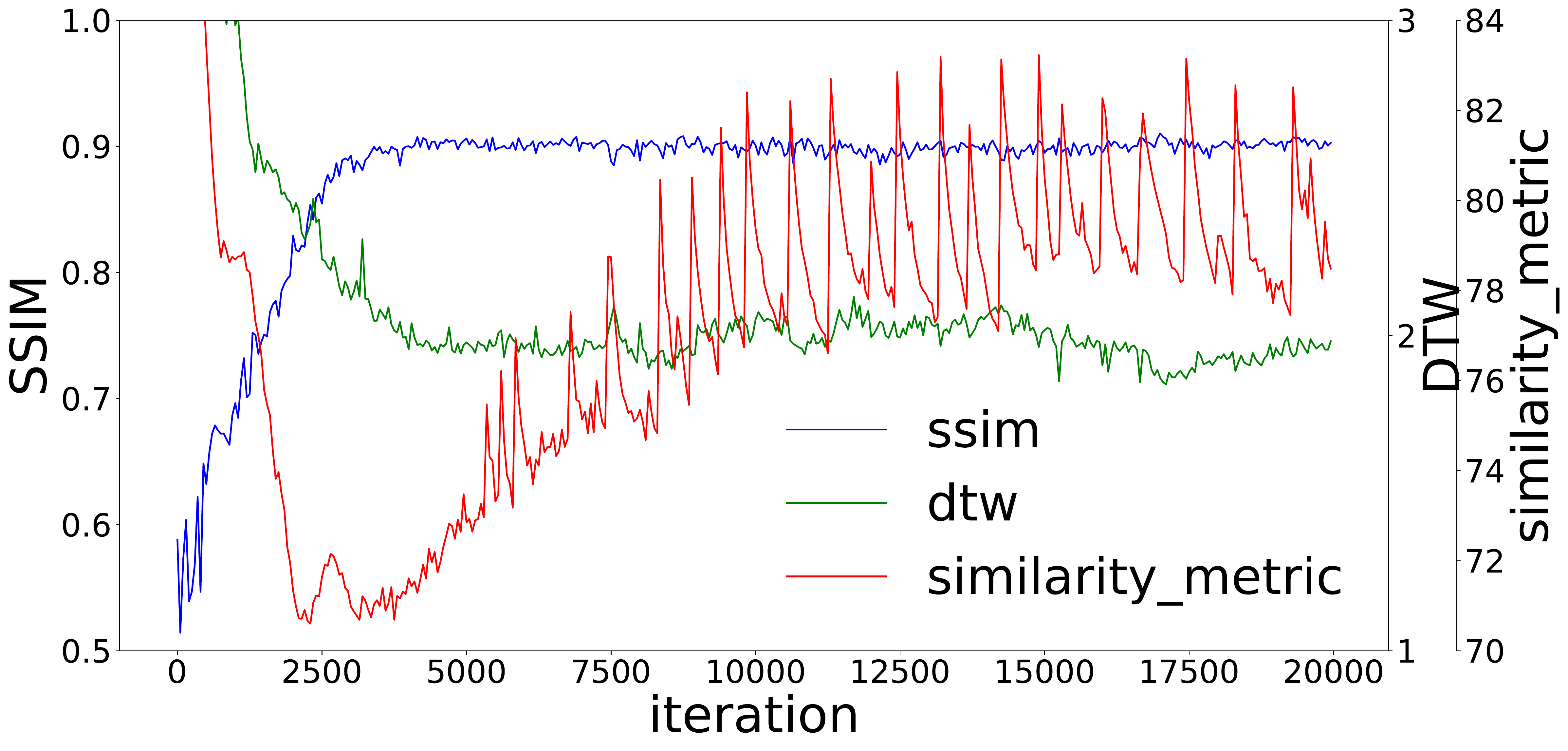}
        \caption{Reconstruction of a test set during training}
        \label{fig:test-hist}
    \end{subfigure}
    \caption{Successfully training a model with $\beta$-250, $\gamma$-100 and $10$ latent dimensions, using the Adam optimizer~\cite{DBLP:journals/corr/Ruder16}}
    \label{fig:training-good}
\end{figure}

Having introduced objective similarity metrics, a calibration exercise helps in understanding their overall relationship with the model. It also gives the advantage of observing training progress through measures that are more intuitive than the training objective. A natural first step is to relate objective metrics with their subjective counterpart, the VAE/GAN similarity metric.  Denoting reconstruction of sample(s) $x$ as $\bar{x} = Dec(Enc(x))$, one notes that $\mathcal{L}_{Dis_l}$ is calculated between $x$ and $\bar{x} $ on each training batch, as a measure of sample reconstruction in the training objective. Extending this idea, upon measuring the similarity between $x$ and $\bar{x}$ on a separate test batch, using $\mathcal{L}_{Dis_l}$, DTW and SSIM, one arrives at a calibrated, generalized, and objective measure of the model's reconstruction fitness.

\begin{figure}[b!]
    \begin{subfigure}{\linewidth}
        \centering
        \includegraphics[width=0.8\linewidth]{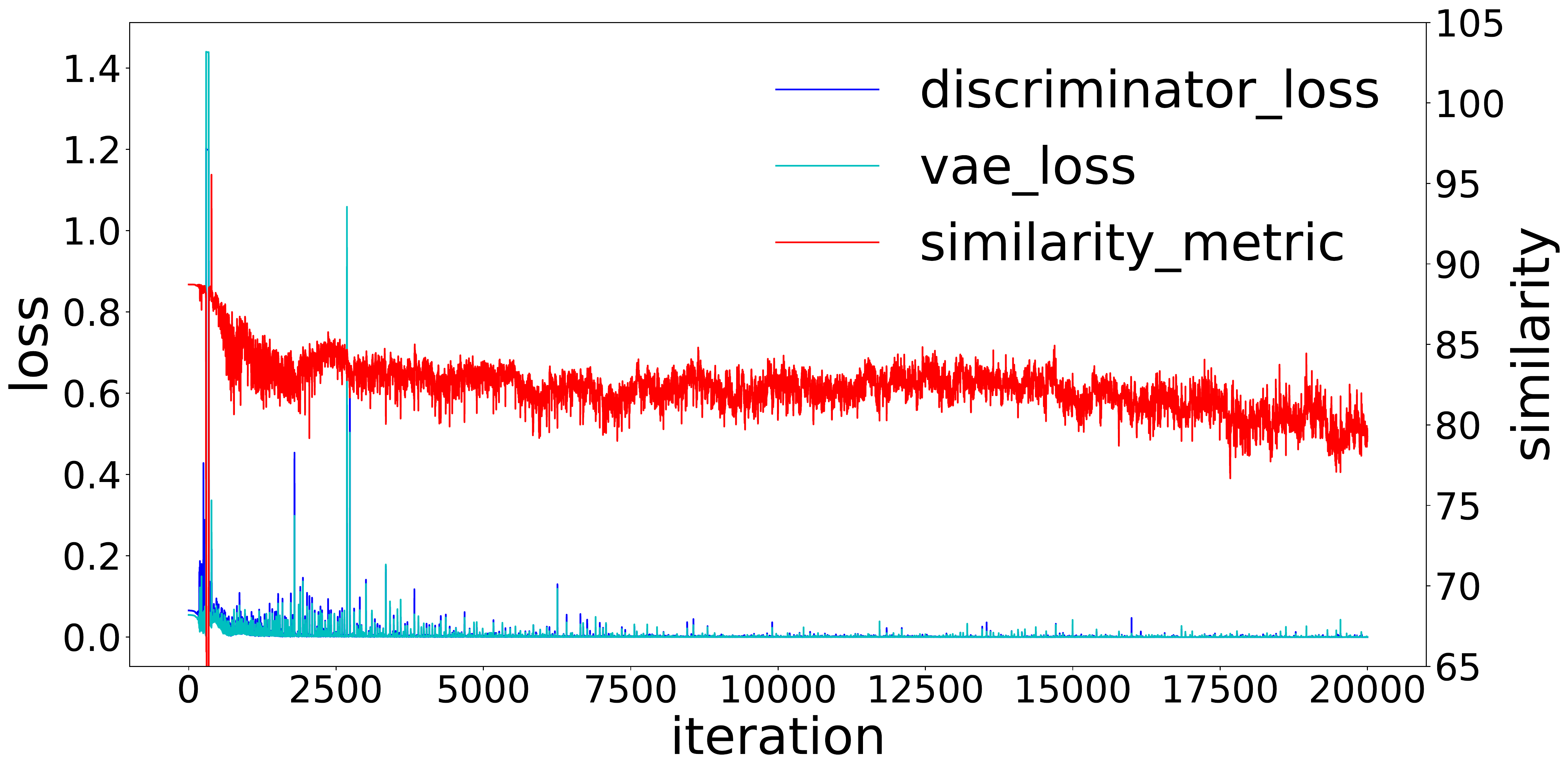}
        \caption{Training history}
        \label{fig:training-hist-bad}
    \end{subfigure}
    \begin{subfigure}{\linewidth}
        \centering
        \includegraphics[width=0.8\linewidth]{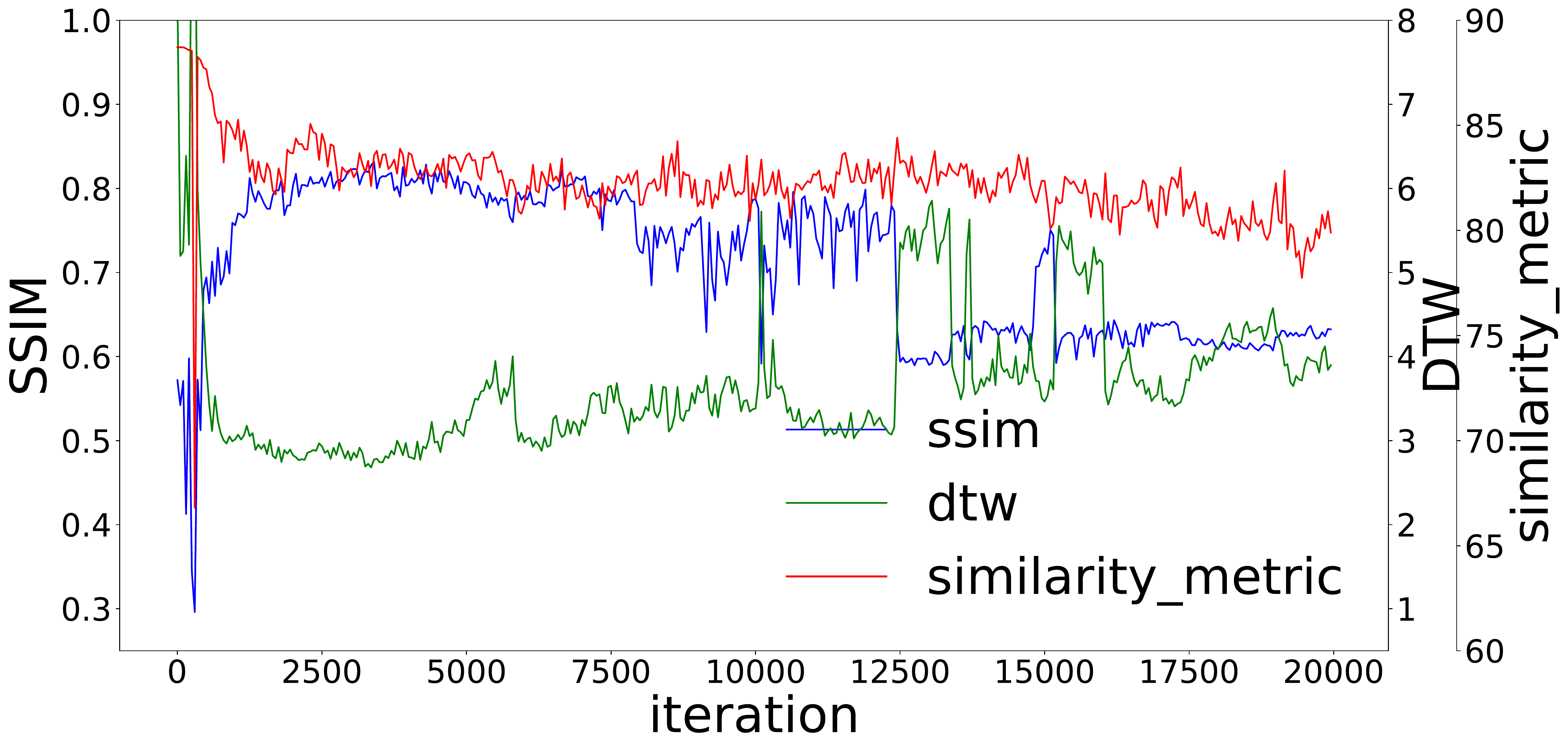}
        \caption{Reconstruction of a test set during training}
        \label{fig:test-hist-bad}
    \end{subfigure}
    \caption{Unsuccessfully training a model with $\beta$-250, $\gamma$-100 and $10$ latent dimensions, using the RMSProp optimizer~\cite{DBLP:journals/corr/Ruder16}}
    \label{fig:training-bad}
\end{figure}

As prescribed in \cite{DBLP:conf/icml/LarsenSLW16}, each training iteration consists of performing mini-batch gradient descent sequentially on the encoder ($\mathcal{L}_{Dis_l} + \beta \mathcal{L}_{prior}$), the decoder ($\gamma \mathcal{L}_{Dis_l} - \mathcal{L}_{GAN}$), and the discriminator ($\mathcal{L}_{GAN}$), averaged over 5 mini-batches.
An example of a successful training regime can be seen in Fig~\ref{fig:training-good}. Convergence trends can be clearly spotted for the VAE/GAN similarity metric (Fig~\ref{fig:training-hist}), which converges around 10k iterations. In fact, there is no observable gain in average test set reconstruction SSIM (Fig~\ref{fig:test-hist}) beyond $\sim0.9$ after 3k iterations. DTW seems to take a longer time to settle at $\sim2$, pointing to higher sensitivity than SSIM. The 'lag' between metrics in showing convergence is understandable, given their individual capabilities. However, the combined picture they present increases confidence in training convergence.

In contrast, an unsuccessful training regime is shown in Fig~\ref{fig:training-bad}. Training objectives (Fig~\ref{fig:training-hist-bad}) present a mixed picture, with the VAE/GAN similarity metric on a slow downward trend without the oscillations found in the successful case. Loss function trends are largely similar to those in the success case. The collapse in SSIM to $\sim0.6$ and DTW to $\sim4$ (Fig~\ref{fig:test-hist-bad}), however, makes it clear that there is indeed no convergence in training. While such a clean indication is not always available, measuring reconstruction fitness using objective metrics does seem to give more intuitive feedback on the training process than the raw training objective. It is also clear that a major challenge lies in developing the notion of a good similarity score, requiring careful metric calibration on the dataset. In our case, we visually inspected the reconstruction of 15 randomly drawn samples from the test set to develop a notion that SSIM $> 0.9$ is healthy (for example, Fig~\ref{fig:stop-tmpl-to-real}c and Fig~\ref{fig:stop-tmpl-to-real}d).\looseness=-1

\subsection{Using metrics to evaluate generator models}
\begin{figure}[h]
    \centering
    \includegraphics[width=0.98\linewidth,trim={0 0 0 8},clip]{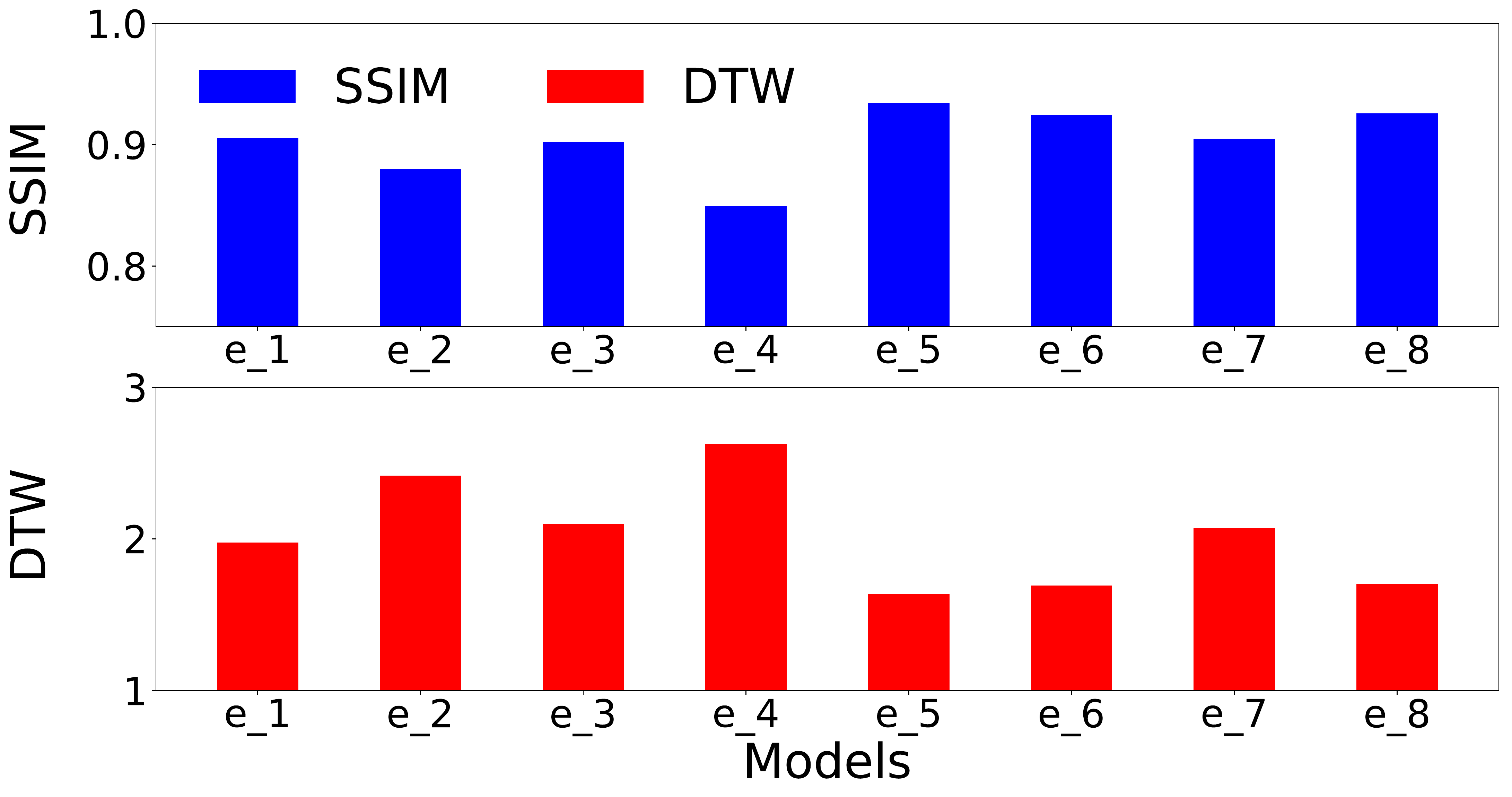}
    \caption{Average (top) SSIM and (bottom) DTW between the test set and its reconstruction, for models in Table \ref{table:models}  }
    \label{fig:summary-reconstruction}
\end{figure}

Consider, for example, the task of evaluating models in Table \ref{table:models}. As seen in Fig~\ref{fig:summary-reconstruction}, SSIM and DTW show quantitative differences between models' reconstruction fitness. The model $e_5$, with highest average reconstruction SSIM and lowest average reconstruction DTW, best reconstructs the test set. While increasing the number of latent dimensions from 5 to 10 seems to increase reconstruction quality, more model samples are necessary to draw conclusions on the influence of $\beta$. We have thus shown a metric driven technique, of reasonable cost, that provides direct and intuitive feedback on generator quality both during and after training.

\begin{table}[h]
\centering
\caption{Explored model configurations, $\gamma$=100 in all cases}
 \begin{tabular}{|l c r|}
 \hline Name & Latent dims & $\beta$ \\ [0.5ex] 
 \hline $e_1$ & 5 & 1\\
 \hline $e_2$ & 5 & 10\\
 \hline $e_3$ & 5 & 50\\
 \hline $e_4$ & 5 & 250\\
 \hline $e_5$ & 10 & 1\\
 \hline $e_6$ & 10 & 10\\
 \hline $e_7$ & 10 & 50\\
 \hline $e_8$ & 10 & 250\\ [1ex]
 \hline
\end{tabular}
\label{table:models}
\vspace{-4mm}
\end{table}

%%%%%%%%%%%%%%%%%%%%%%%%%%%%%%%%%%%%%%%%%%%%%%%%%%%%%%%%%%%%%%%%%%%%%%%%%%
\section{Metric guided stimulus generation}\label{sec:generation}

\begin{figure}[t]
    \centering
    \includegraphics[width=0.98\linewidth]{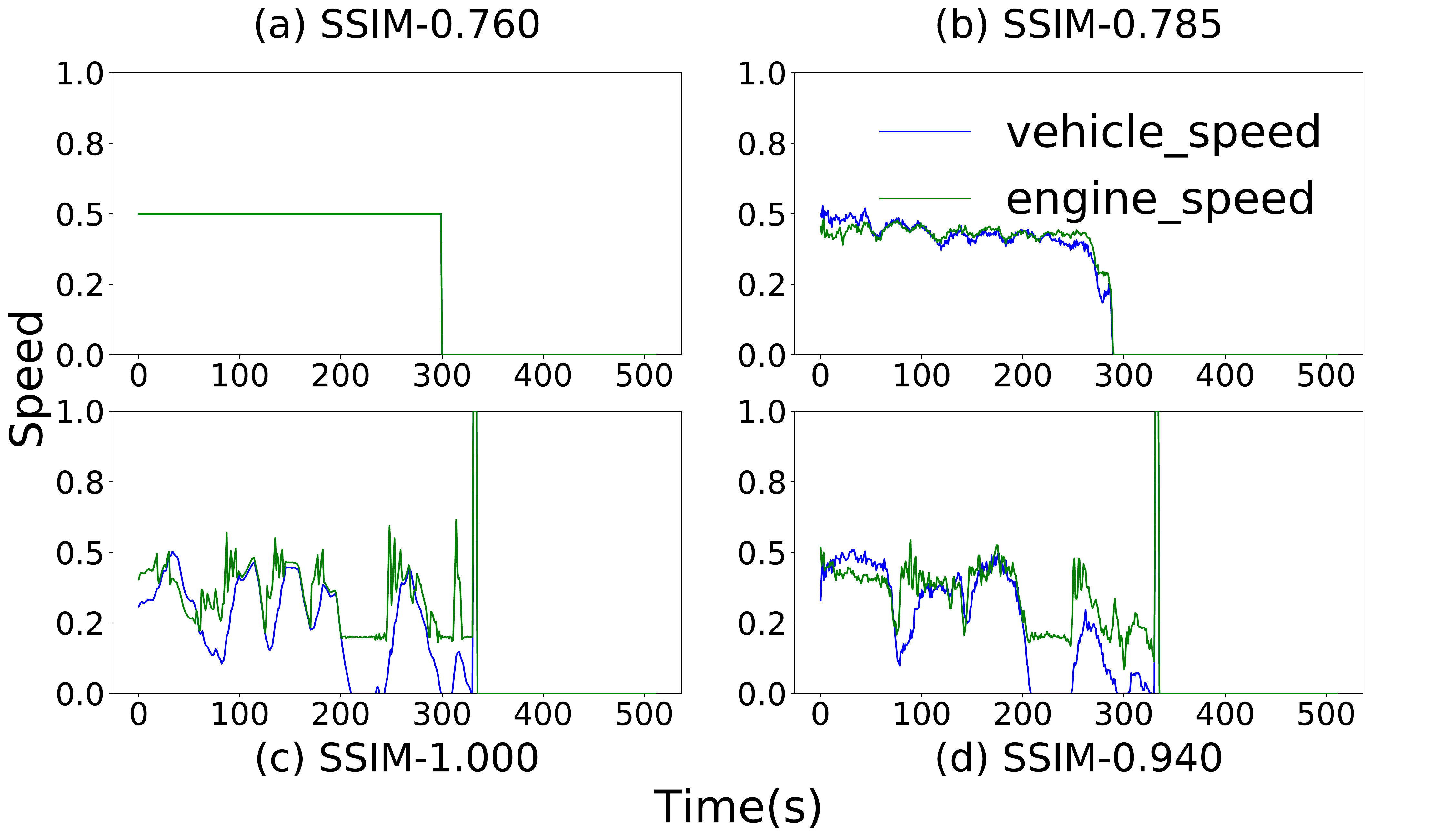}
    %reduce space after caption
    \setlength{\belowcaptionskip}{-10pt}
    \caption{(a) Interpolation source $x_1$ (b) and its reconstruction, (c) interpolation target $x_2$, and (d) its reconstruction . All SSIM values are measured against $x_2$. Speed values rescaled to  [0,1]}
    \label{fig:stop-tmpl-to-real}
\end{figure}

The notion of similarity measurement is now extended to the original objective of helping testers generate stimuli in a controllable fashion.\looseness=-1

\subsection{Metric guided interpolation}
Consider the case of testing SUT behavior when the vehicle comes to a halt. The tester could come up with a template $x_1$ (Fig~\ref{fig:stop-tmpl-to-real}a) as a reference. The template, while capturing the basic idea of the vehicle stopping, is not very realistic. To compensate, the tester could turn to another reference -- a recorded signal showing a similar characteristic (Fig~\ref{fig:stop-tmpl-to-real}c). Generating plausible intermediate sequences between references $x_1$  and $x_2$, with controlled introduction of elements of reality, is a well-structured, realistic, test of stopping behavior.\looseness=-1

In any GAN model, the latent space $z$ encodes a representation of the data distribution. Given $x_1$ and $x_2$, it has been demonstrated (most notably in \cite{DBLP:journals/corr/RadfordMC15}) that linear interpolation between $z_1 = Enc(x_1)$ and $z_2 = Enc(x_2)$, yields points ${z_l}$, which, upon decoding $\bar{x_l} = Dec(z_l)~\forall l$, represents a smooth transition between $x_1$ and $x_2$ in $x$ space. Let us now examine this for the case of constructing intermediate sequences between chosen references $x_1$  and $x_2$. One immediately noticeable advantage is that even a simplistic stop template is rendered, upon reconstruction, into a much more plausible form (Fig~\ref{fig:stop-tmpl-to-real} a and c). The model is therefore a powerful alternative to classical function approximations. While a set of intermediate samples can be readily generated by naive linear interpolation, its plausibility as valid stimuli is vastly enhanced if generated samples show a smooth change in SSIM. Noting that $Decoder$ and $SSIM$ are continuous in their respective domains, it follows from the intermediate value theorem that an iterative search along the straight line between $z_1$ and $z_2$ gives a set of points which, when decoded, represents an arbitrarily smooth increase in SSIM. The granularity of the search is decided by the sampling ratio $s$.\looseness=-1

\begin{algorithm}
\SetKwInOut{Input}{Input}
\SetKwInOut{Output}{Output}
\SetKwBlock{Repeat}{repeat}{}
\begin{small}

    \Input{Start position $z_0$, end position $z^*$, n.o. waypoints $no\_wps$, model $m$, metric $f_m$, sampling ratio $s$}
    \Output{Intermediate latent space positions $zs$}
    
    \vspace{5pt}
    
    %\begin{multicols}{2}
    $x^* \leftarrow m.\text{generate}(z^*)$ \\
    $start\_metric \leftarrow f_m(m,z_0,x^*)$ \\
	$stop\_metric \leftarrow f_m(m,z^*,x^*)$ \\
	$wps \leftarrow \text{interp}(start\_metric, stop\_metric, no\_wps)$ \# interpolation in metric space \label{line:metric_interp} \\
	\vspace{5pt}
	$zs \leftarrow \text{linspace}(z_0, z^*, no\_wps)$ \# initialize with linspace \\
    $num\_steps \leftarrow s \cdot |waypoints|$ \\
    \For {$i\in[num\_steps]$}{
        $pos \leftarrow z_0 + \frac{i}{num\_steps}(z^*-z_0)$ \\
        $m_{pos} \leftarrow f_m(m, pos, z^*)$ \\
        $wp\_ind \leftarrow \text{argmin}_{i \in [|wps|]} (|wps[i] - m_{pos}|)$ \# find closest waypoint in metric space \\
        $m_z \leftarrow f_m(m, zs[wp\_ind], z^*)$ \\
        \If{$m_{pos} < m_z$}{
            $zs[wp\_ind] \leftarrow pos$
        }
    }
    %\end{multicols}
\end{small}
\caption{MLERP}
\label{algorithm:mlerp}
\end{algorithm}

\begin{figure}[h]
    \begin{subfigure}{\linewidth}
        \centering    
        \includegraphics[width=0.8\linewidth]{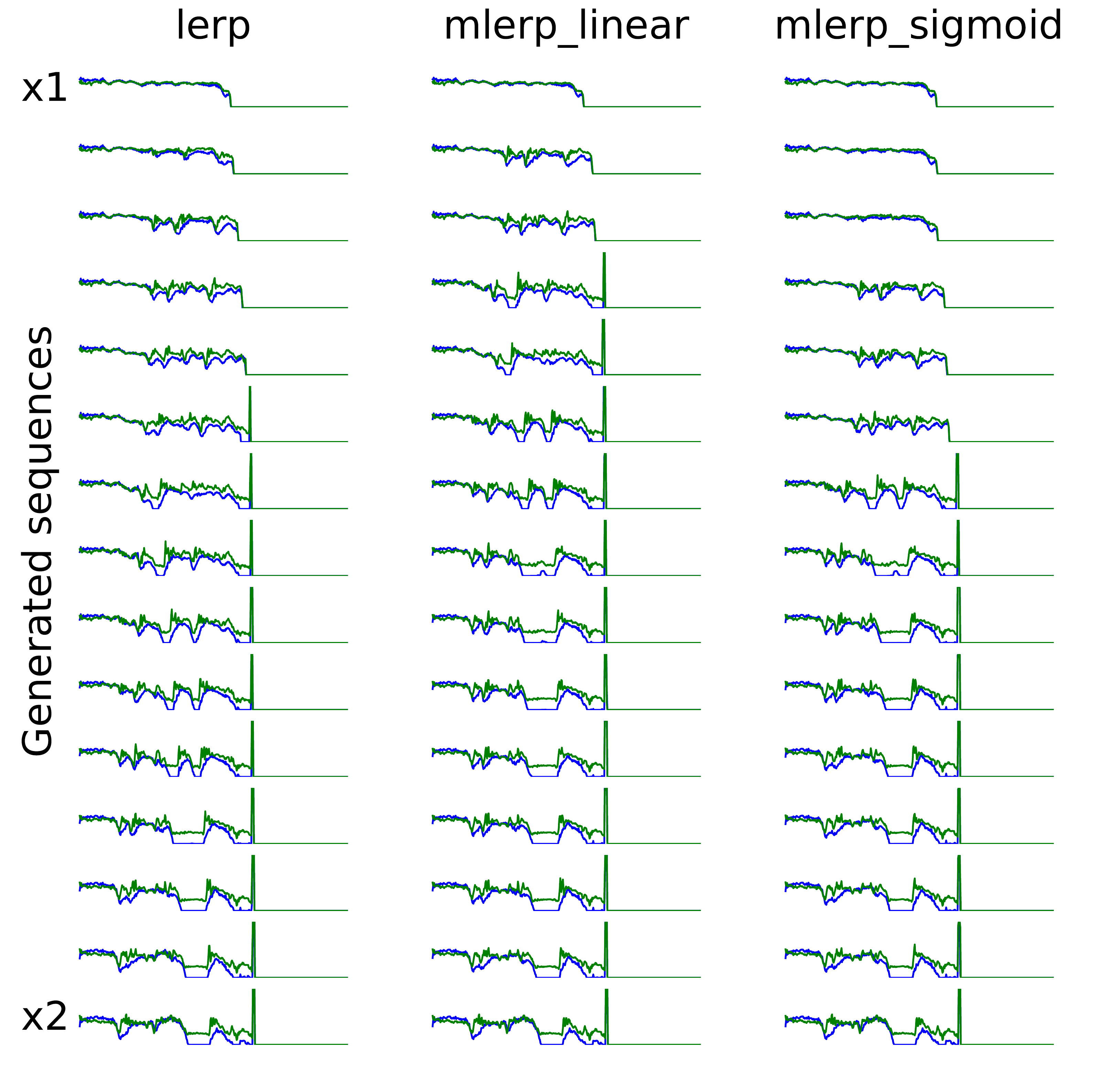}
        \caption{Generated samples}
        \label{fig:stop-lerp-mlerp}
    \end{subfigure}
    \begin{subfigure}{\linewidth}
        \centering
        \includegraphics[width=0.9\linewidth]{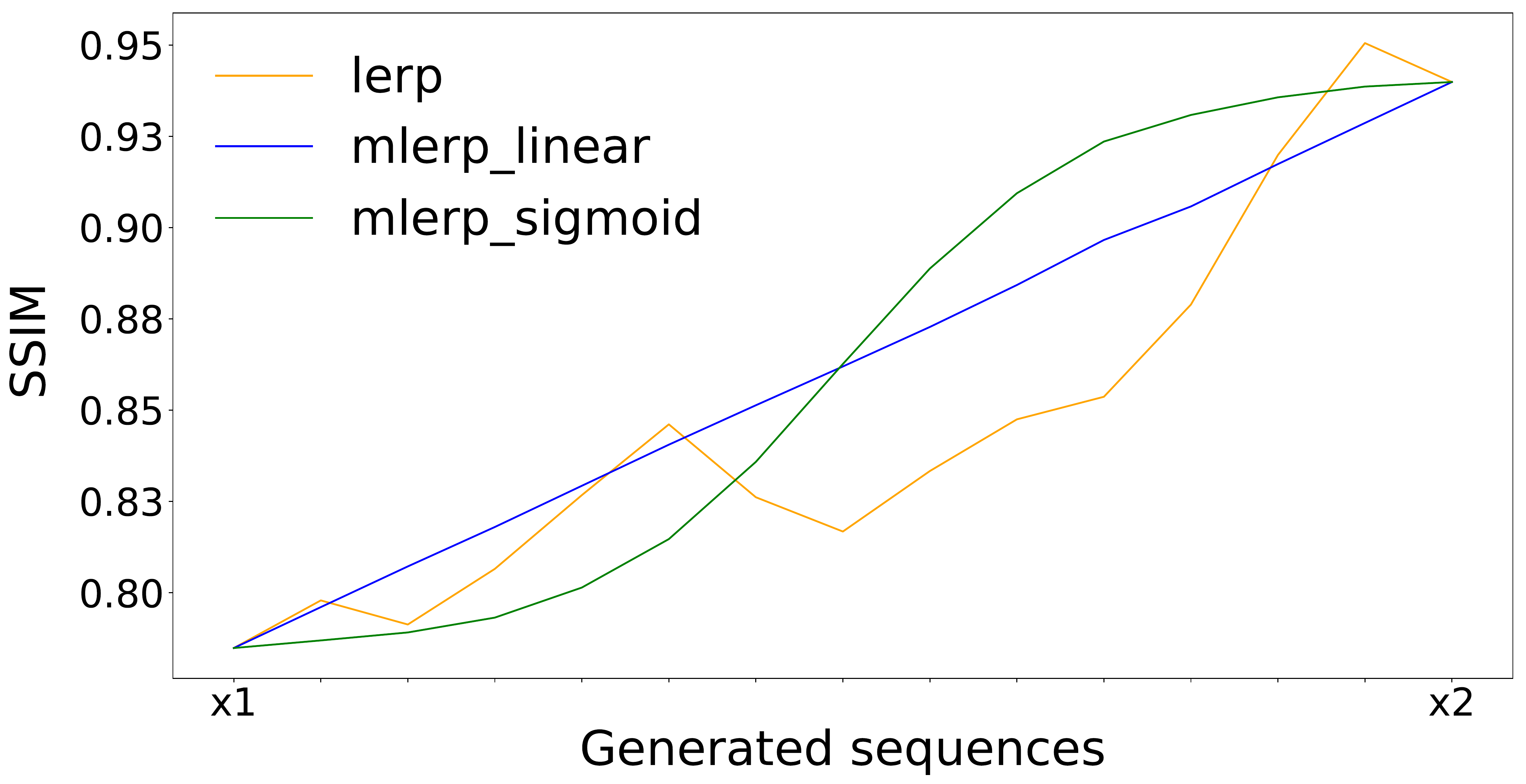}
        \caption{SSIM of generated samples}
        \label{fig:stop-tmpl2real-cmp}
    \end{subfigure}
    %reduce space after caption
    \setlength{\belowcaptionskip}{-10pt}
    \caption{Generating intermediate samples between  template ($x_1$) and real-world ($x_2$) stop sequences with $e_5$ using naive linear interpolation (\textit{lerp}), MLERP with linear increase in SSIM $s$=15 (\textit{mlerp\_linear}), and MLERP with sigmoid increase in SSIM and $s$=30 (\textit{mlerp\_sigmoid})}
\end{figure}

To achieve this, we propose the metric based linear interpolation algorithm MLERP (Algorithm \ref{algorithm:mlerp}) which interpolates in \textit{metric} space, as opposed to the latent. Given some desired division of the metric space (Algorithm~\ref{algorithm:mlerp} -- step \ref{line:metric_interp}), MLERP will sample points along the straight line between $z_1$ and $z_2$, while keeping track of the latent positions which correspond best to the metric space division. The metric $f_m$, sampling ratio $s$, number of waypoints $no\_wps$, and metric space division $wps$ are all tunable parameters. MLERP can be applied for generating a set of intermediate sequences whose SSIM changes smoothly between $x_1$ and $x_2$. Observing the generated sequences themselves (Fig~\ref{fig:stop-lerp-mlerp}) gives an indication of controlled variation. This is best seen in the \textit{metric\_sigmoid} case, where the resemblance with $x_2$ grows gradually because the metric space division $wps$ is set as the logistic sigmoid function. In comparison, \textit{mlerp\_linear} is a case where resemblance grows rapidly, since $wps$ is set to be linear. Fig~\ref{fig:stop-tmpl2real-cmp} clearly shows that naive linear interpolation results in irregular increase in SSIM. MLERP, on the other hand, always guarantees a controllably smooth increase in SSIM.\looseness=-1

\begin{figure}[t]
    \begin{subfigure}{\linewidth}
        \centering
        \includegraphics[width=0.8\linewidth]{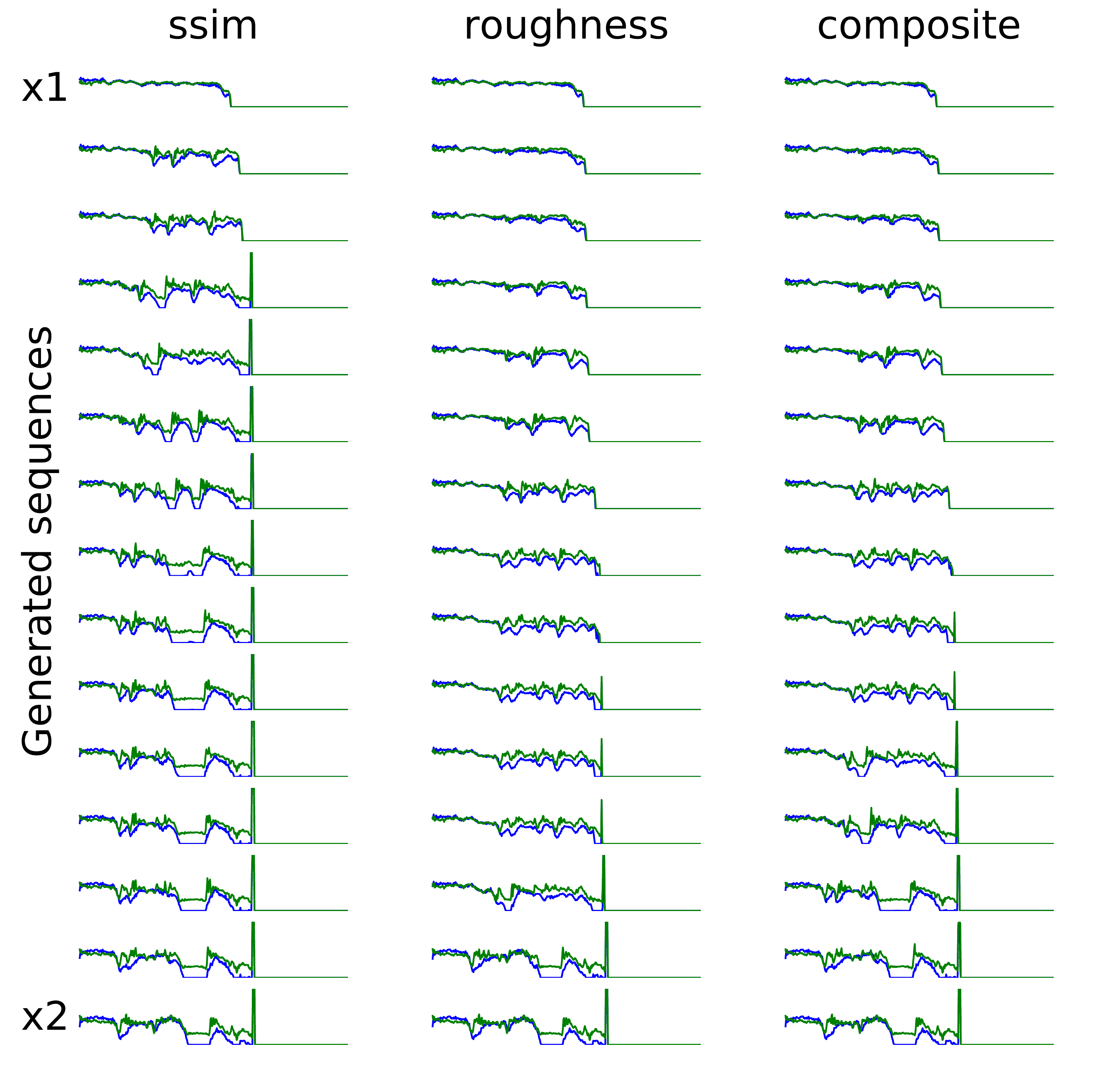}
        \caption{Generated samples}
    \label{fig:mlerp-composite-seq}
    \end{subfigure}
    \begin{subfigure}{\linewidth}
        \centering
        \includegraphics[width=0.9\linewidth]{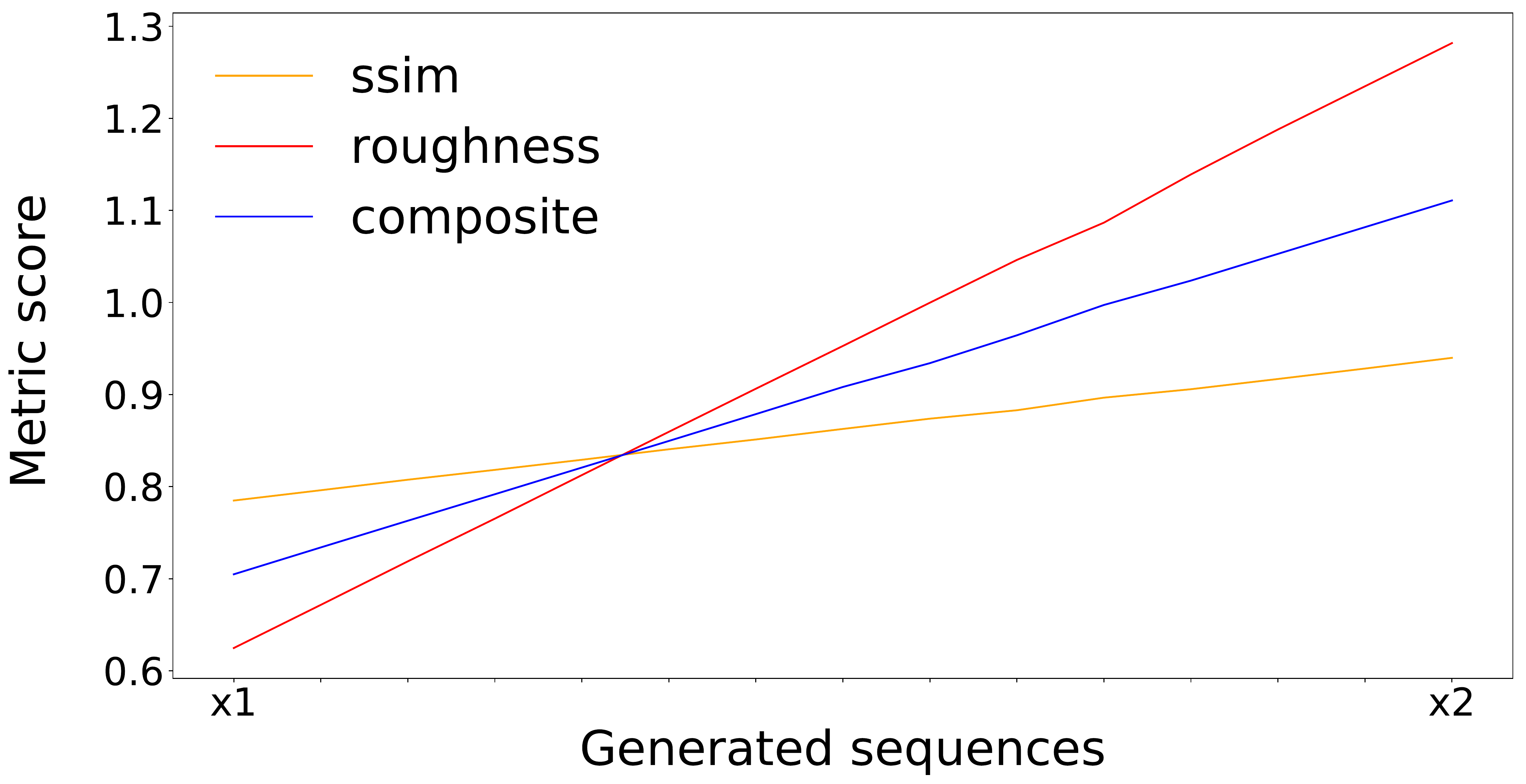}
        \caption{Metric scores of generated samples}
    \label{fig:mlerp-composite}
    \end{subfigure}
    \caption{Generating intermediate samples between  template ($x_1$) and real-world ($x_2$) stop sequences, with $e_5$, using MLERP with SSIM $s$=15 (\textit{ssim}) , $R$ $s$=100 (\textit{roughness}), and $M_{sr}$ $\kappa$=0.5, $s$=100 (\textit{composite})}
\end{figure}

To further illustrate finely controlled generation, consider the composite metric $M_{sr}$ (Eq \ref{eq:msr}). Here SSIM is combined with $R$, which is a simple measure of roughness/smoothness of a given sequence $b$, relative to the reference sequence $a$, calculated as a ratio of average absolute change in sample values. The effect of $R$ is best seen in the \textit{roughness} column in Fig~\ref{fig:mlerp-composite-seq}, where the smoothness of generated sequences steadily reduces, when compared to the \textit{ssim} column, where sequences, guided by SSIM, steadily increase resemblance. Sequences generated using composite metric $M_{sr}$, as seen in the \textit{composite} column in Fig~\ref{fig:mlerp-composite-seq}, shows how a compromise can be struck between these two metrics. Metric variation is best seen in Fig~\ref{fig:mlerp-composite}, where metric interpolation using $M_{sr}$ strikes a balance between MLERP using SSIM and $R$ individually, usually at the cost of a higher $s$.

Using such a set of metric guided synthetic stimuli during a test run not only ensures that test inputs are \textit{verifiably realistic}, but also that \textit{variations are controllable}. These are the twin objectives that testers aim for, but find it difficult to achieve in current practice. While this is certainly beneficial compared to the original choices of crafted and real-world stimuli, realizing full potential of this technique, depends upon choosing the right metric, which is not always trivial. Compared to naive interpolation, metric-search incurs an additional cost in sampling, which is decided by the sampling ratio $s$. Such a cost can be offset when stimulus generation happens less frequently than test case execution and when it is, in the best case, a one-time activity.\looseness=-1
\setlength{\abovedisplayskip}{3pt}
\setlength{\belowdisplayskip}{3pt}
\begin{gather}
    M_{sr}(x_2,\bar{x}) = \kappa\,SSIM(x_2,\bar{x}) + (1-\kappa)R(x_2,\bar{x}) \label{eq:msr}\\ 
    R(a,b) = \frac{\sum_{i=0}^{N-1}|b_{i+1} - b_{i}|}{\sum_{i=0}^{N-1}|a_{i+1} - a_{i}|}
\end{gather} 

\subsection{Metric guided neighborhood search}
\begin{figure}[h]
    \centering
    \begin{subfigure}{\linewidth}
        \includegraphics[width=0.9\linewidth]{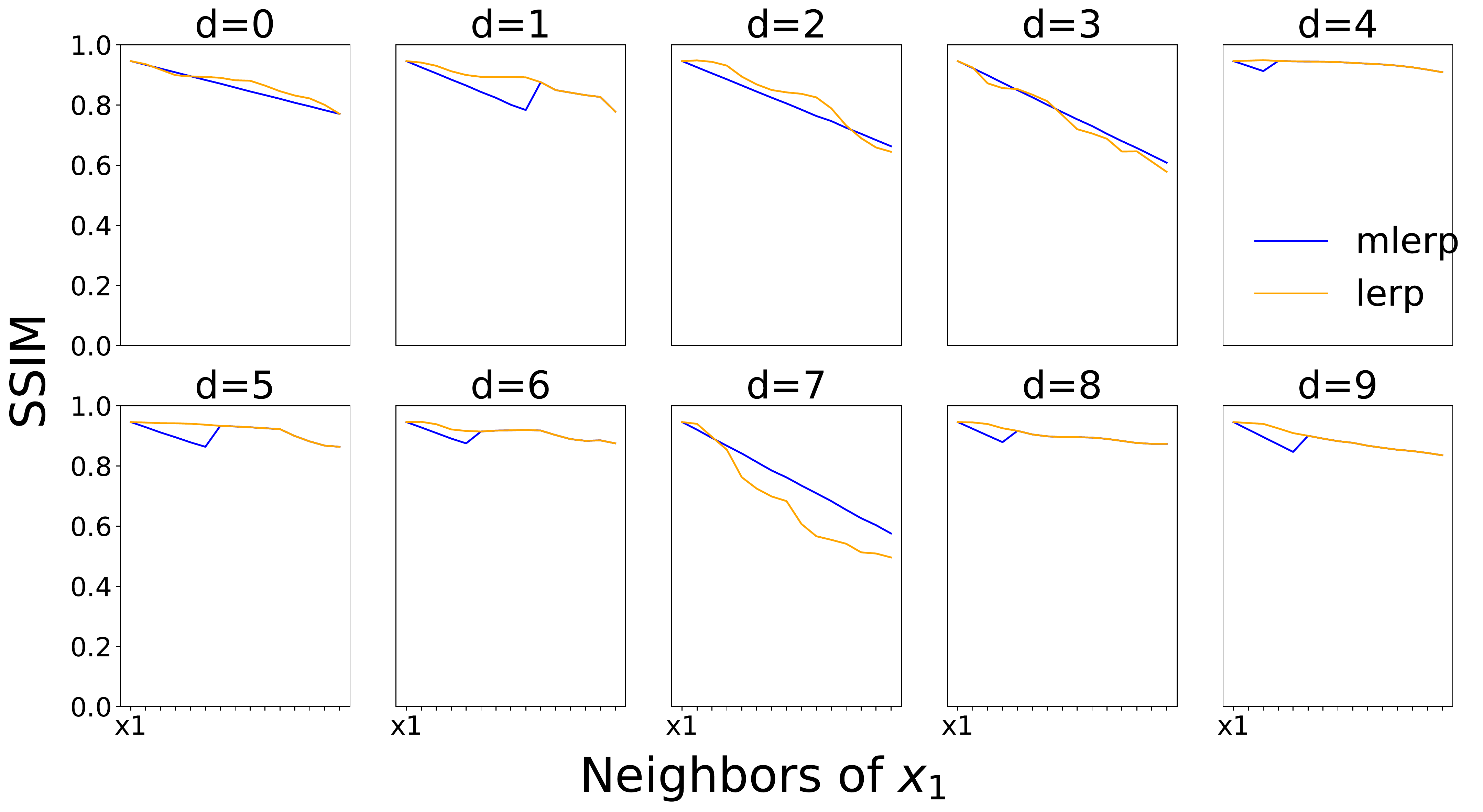}
        \caption{Neighborhood search using $e_5$ ($\beta = 1$)}
        \label{fig:stop-neighbors-cmp-a}
    \end{subfigure}
    \begin{subfigure}{\linewidth}
        \includegraphics[width=0.9\linewidth]{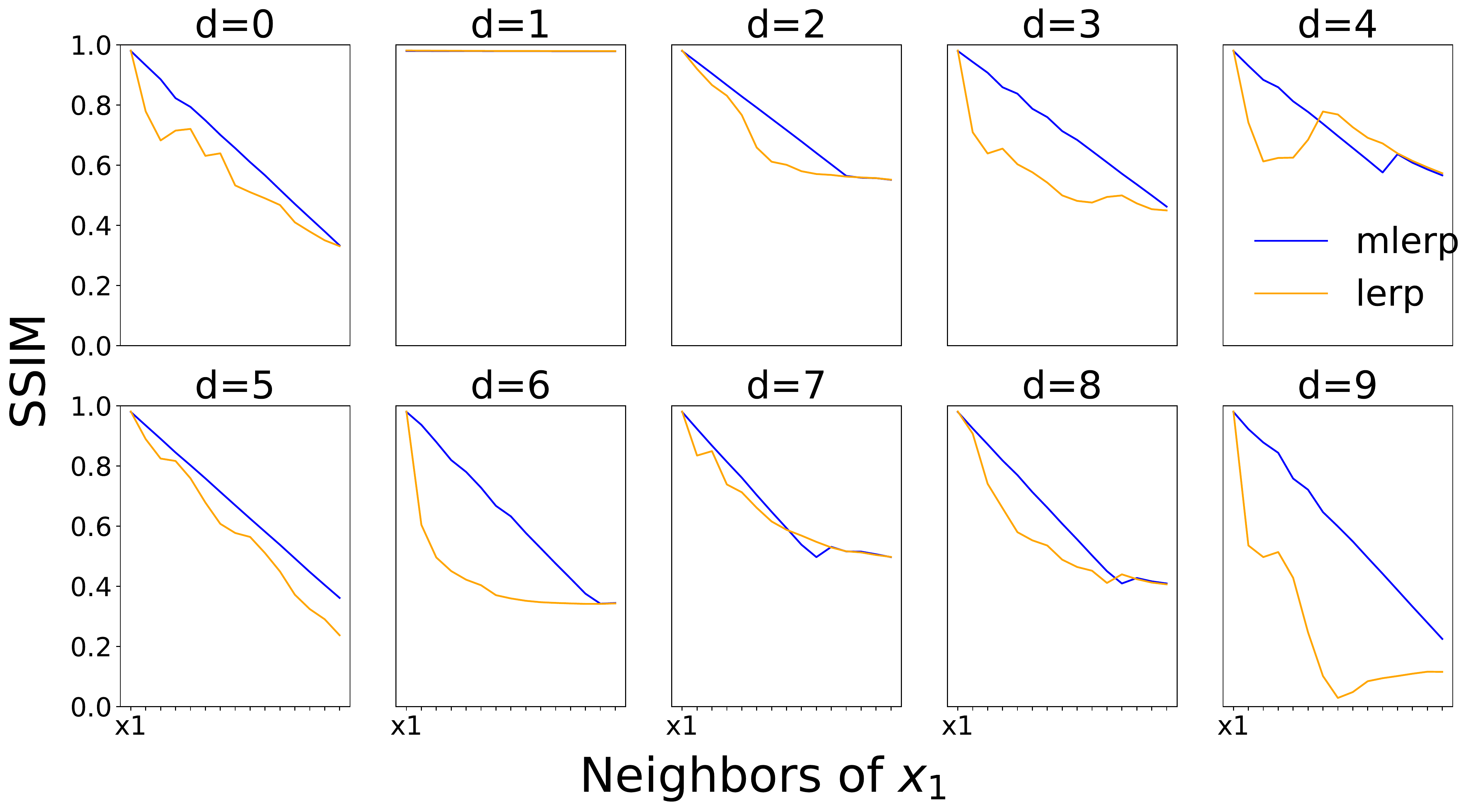}
        \caption{Neighborhood search using $e_8$ ($\beta = 250$)}
        \label{fig:stop-neighbors-cmp-b}
    \end{subfigure}
    \setlength{\belowcaptionskip}{-10pt}
    \caption{Comparison of SSIM variations between neighborhood search using metric (mlerp) and naive (lerp) linear interpolation, along each latent space dimension $d$, between $z_1$ and $z_1[d] += 2$. Metric search was conducted with $s=30$.}
    \label{fig:stop-neighbors-cmp}
\end{figure}

Given a stimulus sequence $x_1$ and its latent code $z_1$, we know there exist samples of similar structure with latent codes in the neighborhood of $z_1$. Metric driven sampling, solely along latent space axes, shows potential in identifying such neighbors, allowing testers to perform fine adjustments to $x_1$. As seen in Fig~\ref{fig:stop-neighbors-cmp-a}, with $\beta=1$, MLERP between $z_1$ and $z_1[d]+=2$, along each latent space dimension $d$, is smoother than naive interpolation in 3 out of 10 cases. To keep search costs low, $s$ is fixed to $30$. Increased disentanglement ($\beta=250$) shows larger and smoother change (Fig~\ref{fig:stop-neighbors-cmp-b}) in SSIM, with same $s$, along more latent space dimensions. Examining variations in SSIM, along latent space axes, therefore helps in acquiring a quantitative notion of structure disentanglement. This can also be put to use, helping testers perform well informed adjustments in stimulus structure. By inspecting generated structural neighbors (Fig~\ref{fig:stop-neighbors-e8} in the appendix) and their SSIM with $x_1$, testers can select the right adjustment that fits the test case. This expands the tester tool-kit by providing yet another method of stimulus design.

%%%%%%%%%%%%%%%%%%%%%%%%%%%%%%%%%%%%%%%%%%%%%%%%%%%%%%%%%%%%%%%%%%%%%%%%%
\section{Similarity as plausibility of synthetic stimuli}
\label{sec:similarity}
For generating time series stimuli, in automotive SIL testing, we have demonstrated the notion of \textit{similarity with reference(s)} as a capable and quantitative measure of plausibility. Specifically, plausibility is ensured in three forms:
\begin{itemize}
    \item Measuring reconstruction of a test set, similarity between $x$ and $\bar{x}=Dec(Enc(x))$, as a fitness measure of the model
    \item Guided generation of intermediate sequences between $x_1$ and $x_2$, by imposing a joint similarity relationship with $x_1$ and $x_2$. Controllability is enhanced by crafting $x_1$ and choosing (a well corresponding) $x_2$ from recordings
    \item Guided generation of structural neighbors of a sequence $x_1$, by imposing a similarity relationship with $x_1$
\end{itemize}

Assessing similarity using SSIM and DTW denote objective plausibility, but only in relation to a reference. While this technique obviously cannot judge the plausibility of a non-referenced, arbitrary set of generated samples, it bears no major consequence to SIL stimulus generation. Each test case has an objective to meet and, to control both execution and feedback, testers normally have a stimulus structure in mind, which can act as the reference. Identifying real-world references, however, is a labeling activity, albeit sparse. For widespread practical use of this technique, the extent of sparse labeling that is needed remains to be seen.
%%%%%%%%%%%%%%%%%%%%%%%%%%%%%%%%%%%%%%%%%%%%%%%%%%%%%%%%%%%%%%%%%%%%%%%%%
\section{Conclusions}
\label{sec:conc}
Integrating and testing automotive mechatronic systems is complex, where each mile of testing in the field increases time and cost to market. With a majority of feature growth in vehicles expected to come in the form of software, enhancing the capability of SIL verification is essential. As a step in this direction, we demonstrate a GAN-based framework to generate synthetic test stimuli. By exploiting the fact that testers often need to generate stimuli that follows a certain structure, we apply the VAE/GAN architecture and similarity metric-driven search for controlled generation of realistic synthetic stimuli. This achieves the twin objective of realistic but controllable stimulation, which helps expand the use of SIL testing. Future enhancements are certainly possible in network architecture to improve the quality of generated data and adaptation between templates and recordings. Easing metric design is important in making this technique practically easy to use and non-linear searching would also help reduce MLERP search costs. Additional work is necessary to close practical gaps in integration with a suitable SIL framework and deployment in a continuous integration pipeline. While such enhancements add value, the fundamental technique shows promise in broadening the applicability of SIL testing.

%%%%%%%%%%%%%%%%%%%%%%%%%%%%%%%%%%%%%%%%%%%%%%%%%%%%%%%%%%%%%%%%%%%%%%%%%%%%%%%%
\section{Acknowledgements} % more needed here?
We thank Henrik L\"onn and other colleagues at Volvo, Christian Berger, and Carl-Johan Seger for their helpful feedback.

This work was supported by the Wallenberg Artificial Intelligence, Autonomous Systems and Software Program (WASP), funded by the Knut and Alice Wallenberg Foundation.

\bibliographystyle{ieeetr}
\bibliography{bibliography}

%\clearpage

\appendix
\subsection{Supporting figures}
\begin{figure}[h]
    \centering
    \includegraphics[width=1.0\linewidth]{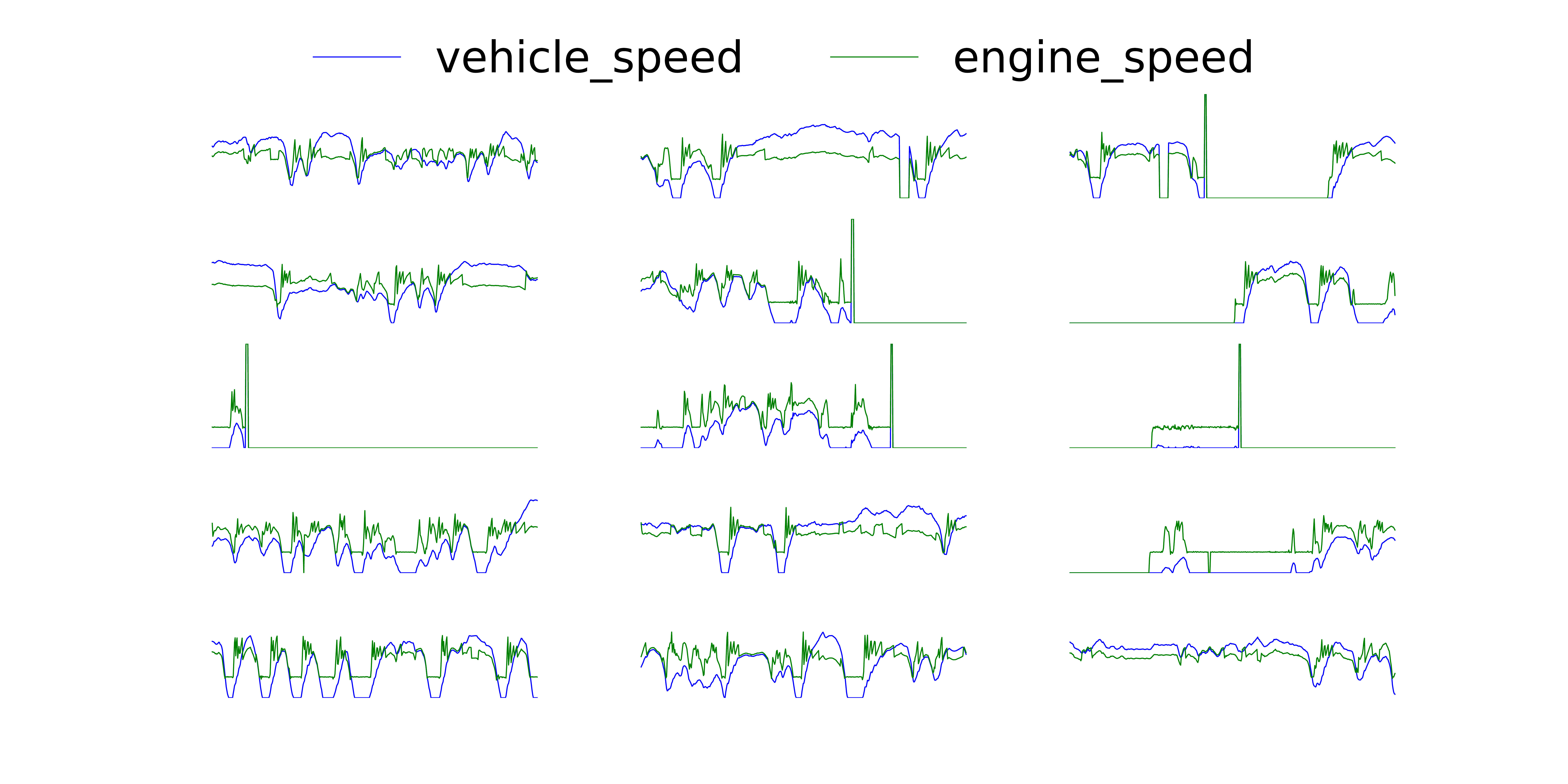}
    \caption{Examples of recorded sequences from the test set (Speed values rescaled to [0,1])}
    \label{fig:test-set}
\end{figure}

\begin{figure}[h]
    \centering
    \includegraphics[width=0.8\linewidth]{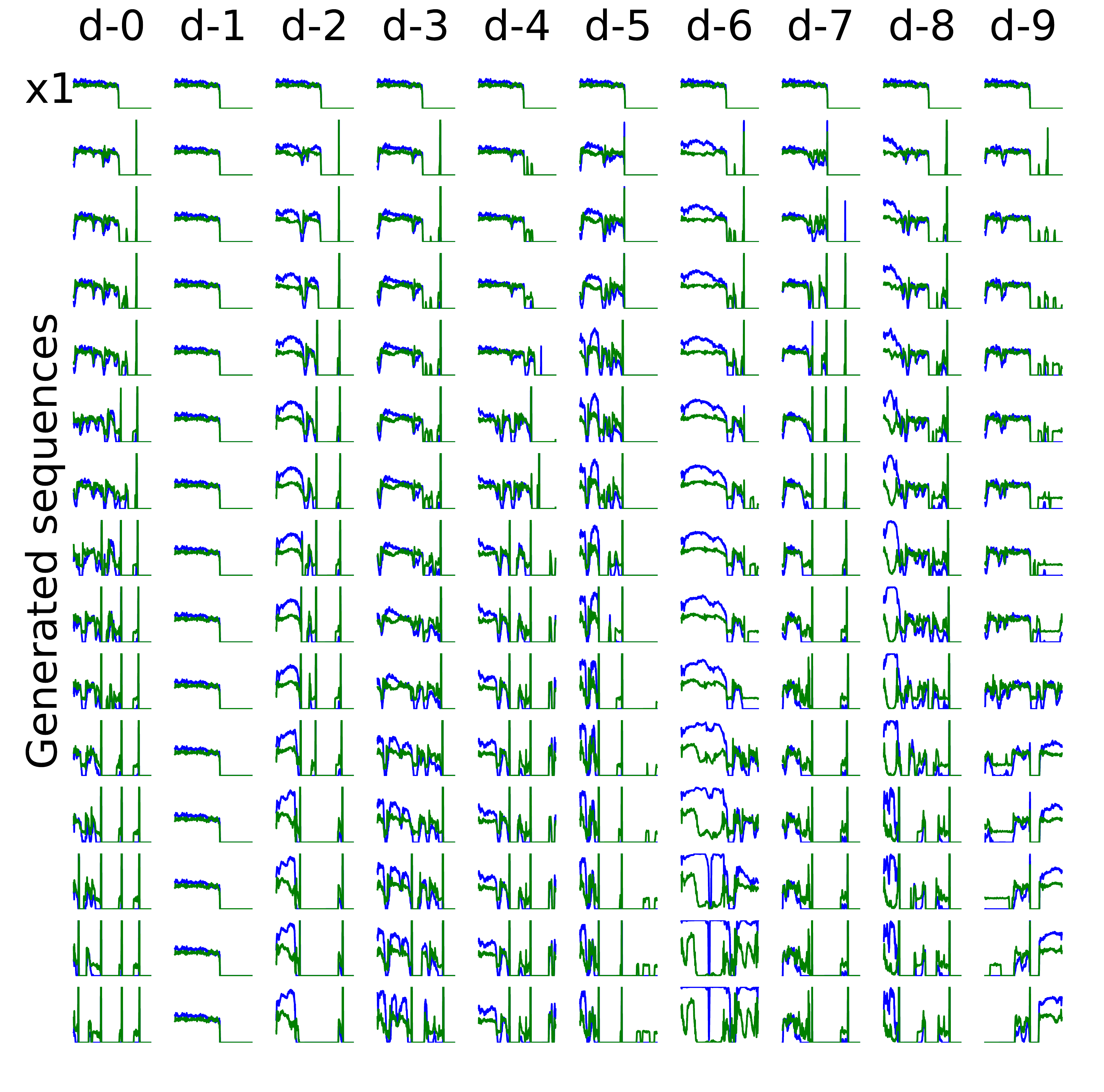}
    \caption{Neighbors of the template stop sequence $x_1$, generated with $e_8$ ($\beta=250$), using metric-guided search. Each column represents interpolation along one latent space dimension $d$ between $z_1$ and $z_1[d] += 2$. Metric search was conducted with $s=20$}
    \label{fig:stop-neighbors-e8}
\end{figure}

\end{document}